\documentclass{article}

\pdfoutput=1

\usepackage{graphicx}
\usepackage{lipsum}
\usepackage[hidelinks]{hyperref}
\usepackage{tabularx}
\usepackage[T1]{fontenc}
\usepackage{calc}
\usepackage[separate-uncertainty=true]{siunitx}
\usepackage{color}
\usepackage[table, dvipsnames]{xcolor}

\usepackage[numbers,sort]{natbib}

\usepackage{subcaption}

\usepackage{bm}

\usepackage{epstopdf}
\usepackage[margin=0.75in]{geometry}

\usepackage{amsfonts}

\usepackage{amsmath}

\usepackage{cleveref}
\usepackage{enumitem}

\usepackage{booktabs}

\title{\textbf{Solving QUBO on the Loihi 2 Neuromorphic Processor}}
\date{\normalsize\today}
\author{
\parbox{\linewidth}{\centering
Alessandro Pierro\textsuperscript{1, 2*},  
Philipp Stratmann\textsuperscript{1,*}, 
Gabriel Andres Fonseca Guerra\textsuperscript{1,*}
Sumedh Risbud\textsuperscript{1}, 
Timothy Shea\textsuperscript{1}, 
Ashish Rao Mangalore\textsuperscript{1, 3},
Andreas Wild\textsuperscript{1}, 
\\%
}
}
\date{\vspace{-0.5\baselineskip}}

\begin{document}

\begin{titlepage}
\thispagestyle{empty}
\pagenumbering{gobble} 
\maketitle

\noindent
\textsuperscript{*} Shared first authors.\\
\textsuperscript{1} Neuromorphic Computing Lab, Intel Labs \\
\textsuperscript{2} Ludwig-Maximilians-Universit\"{a}t M\"{u}nchen \\
\textsuperscript{3} Technische Universit\"{a}t M\"{u}nchen \\

\vspace{\baselineskip}

\setlist[description]{font=\normalfont}
\begin{description}[font=\normalfont,leftmargin=!,labelwidth=\widthof{Number of words Main Text: }]
	\item[Corresponding author:] Andreas Wild\\ \url{andreas.wild@intel.com}
\end{description}

\end{titlepage}

\clearpage
\pagenumbering{arabic}

\section*{Abstract}
In this article, we describe an algorithm for solving Quadratic Unconstrained Binary Optimization problems on the Intel Loihi 2 neuromorphic processor.
The solver is based on a hardware-aware fine-grained parallel simulated annealing algorithm  developed  for Intel's neuromorphic research chip Loihi 2.
Preliminary results show that our approach can generate feasible solutions in as little as \SI{1}{\milli\second} and up to $37x$ more energy efficient compared to two baseline solvers running on a CPU.
These advantages could be especially relevant for size-, weight-, and power-constrained edge computing applications.

\section*{Introduction}

Mathematical optimization (henceforth, simply {\em optimization}) underlies solutions to many problems across industry, science, and society.
The goal is to optimize a cost function over continuous or discrete decision variables of these problems and arrive at an optimal decision.
Many algorithms solve optimization problems by iteratively updating variables connected by sparse, weighted connections. This approach aligns well with the architecture of neuromorphic processors. These chips provide fine-granular parallelism to accelerate computation of objective functions and apply variable updates, they integrate compute with memory to reduce the time and energy cost of data movement, and they support sparse message passing to optimize communication for complex, real-world problems. Inspired by this finding, we have previously applied the Intel Loihi 2 neuromorphic processor to 
two broad optimization problem types, continuous, convex quadratic programming and combinatorial constraint satisfaction, we have shown that the Loihi architecture can solve such polynomial-time and NP-complete problems faster and orders of magnitude more efficiently than state-of-the-art solvers running on CPU and GPU platforms, 
 our results include general QP \citep{mfr24}, unconstrained QP with Lagrangian augmentation \cite{davies_advancing_2021} and constraint satisfaction \cite{FonsecaGuerra2020Stochastic, davies_advancing_2021}.

In this paper, we apply Loihi 2 to the task of solving NP-hard combinatorial problems with discrete variables, specifically quadratic unconstrained binary optimization (QUBO) problems.
QUBO is a problem type that, despite having a simple form, has broad applicability \cite{and14}.
The goal is to identify the binary variable assignment that optimizes a quadratic cost function,
\begin{equation}
\label{eq:qubo}
\min_{\mathbf{x}\in\{0,1\}^n} E(\mathbf{x}) = \min_{\mathbf{x}\in\{0,1\}^n} \mathbf{x}^T \mathbf{Q}\mathbf{x}
\end{equation}
without any constraints\footnote{Optimization problems with constraints can be formulated as QUBO problems by incorporating the constraints into the cost function using Lagrange multipliers.}. Preliminary results presented in this paper, together with the prior work on quadratic programming \cite{mfr24}, demonstrate that Loihi 2 has the potential to solve a wide range of mathematical optimization problems efficiently.

Finding a global optimum of a general QUBO problem is known to be NP-hard \cite{Barahona1982}; however, many applications---especially those operating under latency and energy constraints---are well served by good approximate solutions found by heuristics or stochastic solvers. State-of-the-art solvers for QUBO include variants of Tabu Search and Simulated Annealing \cite{DunningEtAl2018}.
Tabu Search \cite{glover1986future} explores the search space of a problem in an iterative fashion and creates a {\em tabu list}, which filters out prohibited moves based on different criteria. The efficacy of the tabu search algorithm is dependent on a trade-off between speed and memory-usage. A longer tabu list leads to faster approach to a solution at the expense of more memory required to store the list. In prior, unpublished work, we have found that the tabu solver included in the D-Wave Samplers package \cite{dwavesamplers} is the fastest and most optimal open-source CPU solver available and represents the state-of-the-art for non-parallel QUBO solvers.

Simulated Annealing (SA) \cite{optBySimAnn}, in contrast, harnesses Markov chain Monte Carlo (MCMC) sampling to efficiently explore solution spaces. In the classic formulation, the computational complexity of SA is dominated by the computation of a vector-matrix product representing the local cost of each variable update. For that reason, a variety of parallel implementations have been proposed \cite{gre90} and implemented on GPUs \cite{carpenter2010graphics,oshiyama_benchmark_2022,goto_combinatorial_2019}. However, many QUBO formulations of important optimization problems include high levels of unstructured sparsity \cite{glover2022applications}. GPUs, in contrast, have been optimized for dense matrix arithmetic and their dense parallelism may be underutilized and not efficient compared to a dedicated sparse, serial implementation of SA (e.g. \cite{dwavesamplers}).
Second, for large QUBO problems, we observe that SA solver performance on conventional processors tends to be memory-bound due to the need to move matrix chunks in and out of the processor cache.
Several digital application specific integrated circuits (digital ASICs) have been specifically designed and optimized for efficient, parallel SA, including Toshiba's FPGA-based Simulated Bifurcation Machine \cite{tatsumura_fpga-based_2019,goto_combinatorial_2019,tatsumura_scaling_2021,goto_high-performance_2021, kashimata_efficient_2024}, Fujitsu's Digital Annealer \cite{aramon_physics-inspired_2019}, Hitachi's CMOS Annealer \cite{yoshimura2020cmos}, various analogue compute mechanisms \cite{mmb22, jsh23}, and D-Wave's Quantum Annealer \cite{jag11, oot19}. So far, several factors have limited applicability of these platforms: analogue and quantum hardware platforms suffer from noise and impose strict constraints on the structure and size of the QUBO problems that can be mapped on to them. And in general, the specialized design of these ASICs makes them less effective in applications with size, weight, and power constraints, where versatile computing hardware is beneficial to address diverse and evolving computational challenges.

In contrast, neuromorphic solvers for combinatorial optimization problems have been explored for a variety of problem sizes, ranging from small to moderate sizes with tens to hundreds of variables. These have been based on modified Hopfield networks or Boltzmann machines with a combination of search, gradient descent, stochastic noise, or oscillatory dynamics and applied to constraint satisfaction problems, graph problems, or QUBO \cite{FonsecaGuerra2020Stochastic, steidel2018mapcoloring, Corder2018VertexCover, Rahman2017Cognitive, dwo21,  Binas2016Spiking, Mniszewski2019Graph, henke_sampling_2023, Alom2017QUBO, Liang2019Neuromorphic}.
The largest NP-\emph{complete} problems solved on a neuromorphic platform in the literature include: the Latin squares problem up to 400 variables on the Intel Loihi neuromorphic test chip (Loihi 1) \cite{dwo21} and the map coloring problem up to 193 variables on SpiNNaker \cite{fonseca_guerra_using_2017} and Loihi 1 \cite{dwo21, FonsecaGuerra2020Stochastic}.
The largest NP-\emph{hard} problems solved with a neuromorphic processor in prior work were graph partitioning problems up to 30 variables on IBM's TrueNorth chip \cite{Mniszewski2019Graph} and sparse coding problems up to 64 variables on Loihi \cite{Henke2023Sampling}. To our knowledge, no systematic power-performance benchmarking of neuromorphic solvers has been performed against state-of-the-art conventional solvers for such problems, except the results presented in \cite{dwo21, FonsecaGuerra2020Stochastic}.

In this paper, we introduce an approach to solve large, sparse QUBO problems with the Intel Loihi 2 neuromorphic processor. This approach significantly expands the scale of problems and overcomes many challenges with prior proof-of-concept algorithms.
We provide preliminary performance comparisons of the algorithm against two baseline algorithms on CPU. The results show that our hardware-aware, Loihi 2-based simulated annealing algorithm is capable of finding feasible solutions to problems up to $1000$ variables within $1$ ms and requires $37\times$ lower power than the baseline algorithms on CPU.
This paper details the architecture, implementation, and performance of our neuromorphic QUBO solver.
\label{sec:introduction}

\section*{Hardware-aware simulated annealing}
We set out to develop an algorithm inspired by classic simulated annealing (SA) algorithm, that leverages the features and satisfies the constraints of Intel Loihi 2. This endeavor is inspired by the observation that SA matches the unique set of features of neuromorphic computing well, as elaborated in \hyperref[tab:advantages]{table \ref{tab:advantages}}.

\begin{table}[ht!]
\caption{\label{tab:advantages} Loihi 2 provides a unique set of properties in comparison to conventional CPUs and GPUs, which make it benefitial for QUBO heuristics.}
\begin{tabular}{p{0.35\textwidth}|p{0.60\textwidth}} 
    \hline
    \rowcolor{lightgray}{Loihi 2 feature} & {Benefit for QUBO heuristics} \\ \hline
    {Memory-compute integration}  & {Performant data access enables quick \& efficient iterative updates } \\
    {}  & \hspace{0.5cm} {of neuronal states.}\\
    {}  & {Easy scalability since each core computes on its own memory.} \\
    \midrule
    Massive parallelism  & Simultaneously update neuronal states for all variables.  \\
     & Simultaneously check Boltzmann condition for all variables. \\
     \midrule
    Fine-grained parallelism  & Update neurons with different computations in parallel.  \\
    {}  & Apply different computations for different parts of the solution space  \\
    {}  & \hspace{0.5cm} (applied by state-of-the-art optimization packages, not used here). \\
    \midrule
     Sparse communication & Acceleration of the sparse information exchange, communicating, \\
    {}  & \hspace{0.5cm} {e.g., increasingly sparse subset of flips $\Delta x_i$ per iteration.} \\
    \midrule
    Unstructured sparse matrix support & Optimized for real-world workloads with sparse variable interactions\\
    {}  & \hspace{0.5cm} {in $\mathbf{Q}$.} \\
    \midrule
    \bottomrule
\end{tabular}
\end{table}

\subsection*{Conventional simulated annealing}

SA is a a widely used meta-heuristic that has a Boltzmann machine at its core. These are neural networks wherein each neuron $n_i$ encodes the evolution of a {\em binary} variable $x_i$ and computes the change $\Delta E_i$ in the overall energy potentially induced by flipping of $x_i$ ($0\rightarrow 1$ or $1\rightarrow 0$). The Boltzmann machine flips a variable with probability 1 if such a change of state reduces the overall energy, i.e. $\Delta E_i < 0$. If flipping $x_i$ increases the over energy (i.e., $\Delta E_i > 0$), the variable is flipped with probability
\begin{equation}
    \label{eq:boltzmann}
    p \propto exp(-\frac{\Delta E_i}{T})\ .
\end{equation}
Here, the temperature parameter $T$ encodes the level of noise in the stochastic network dynamics and is evolved by the SA algorithm according to a predefined annealing schedule.
While traditional SA checks the Boltzmann condition \hyperref[eq:boltzmann]{equation \eqref{eq:boltzmann}} for all variables sequentially, parallelized versions of SA exist for parallel compute hardware like GPUs \cite{gre90}. A parallel implementation of simulated annealing leads to the issue that if two or more variables fulfill the Boltzmann condition, their simultaneous flipping can substantially worsen the solution due to interaction terms $Q_{ij}$. Fujitsu's parallel Digital Annealer \citep{mtm20}, for example, resolves this issue by using parallelism solely to determine in parallel which of the variables are suitable to fulfill the Boltzmann condition, while flipping only a single one of them.

\subsection*{Simulated annealing by neural dynamics}

\begin{figure}
    \centering
    \includegraphics[width=0.65\linewidth]{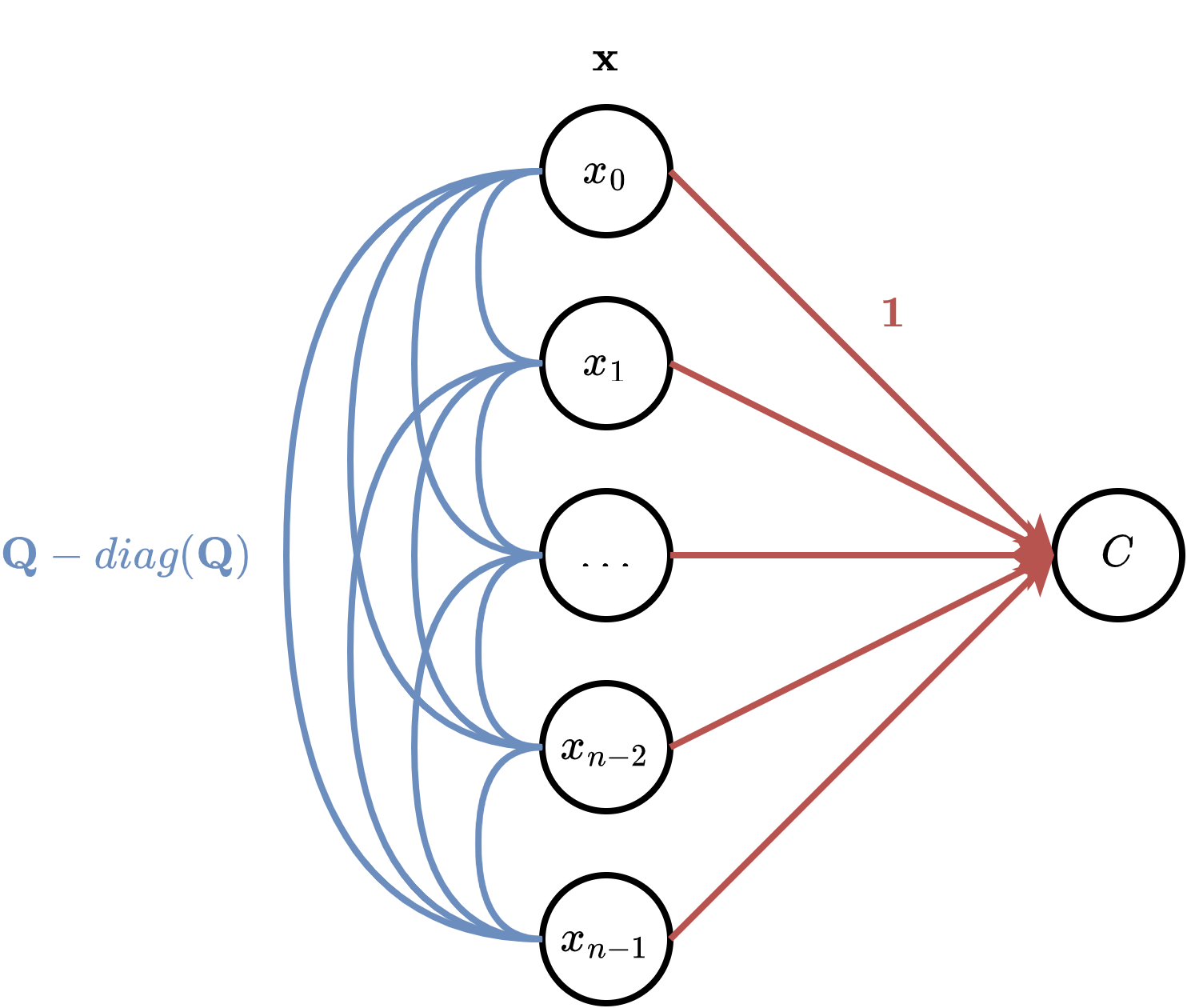}
    \caption{Diagram for the proposed QUBO neural architecture: the variable neurons are recursively connected by a layer of synapses (in blue), which encode the coefficients of the Q matrix, and to the cost integrator by synapses with unit weights (in red).}
    \label{fig:snn-sa}
\end{figure}

The SA framework was used to design a spiking neural network (SNN) architecture which (1) represents a QUBO problem, (2) stores a candidate solution for this problem, and (3) computes the QUBO cost of the candidate solution. Given a variable assignment $\mathbf{x}$, the computation of the cost function can be decomposed as
\begin{align}
    \mathcal{C}(\mathbf{x}) & = \mathbf{x}^T\mathbf{Q}\mathbf{x} \\
    & = \sum_{i,j=0}^{n-1} q_{ij}x_ix_j \\
    & = \sum_{i=0}^{n-1}\left[x_i\left(q_{ii} + \sum_{j\not=i} q_{ij}x_j \right)\right],
    \label{eq:qubo-cost-decomposition}
\end{align}
and we will exploit this last formulation to parallelize the computation.

In the QUBO SNN architecture, outlined in \hyperref[fig:snn-sa]{Figure \ref{fig:snn-sa}}, the variable assignment is maintained by an $n$-dimensional array of \textbf{variable neurons}. At a given step $t$, the $i$-th variable neuron stores the $i$-th binary component of the current candidate solution $x_i^{(t)}$, as well as its value in the previous two steps. These neurons are connected together by a synaptic layer encoding the off-diagonal elements of $\mathbf{Q}$ matrix. Thus, spiking the current binary variable assignment, each of the variable neurons accumulate in the next step $t+1$ the quantity
\begin{align}
    z_i^{(t+1)} & = (\mathbf{Q}-\text{diag}(\mathbf{Q}))_i^T\mathbf{x}^{(t)} \\
    & = \sum_{j\not=i}q_{ij}x_j^{(t)}
\end{align}
which corresponds to the contribution of the mixed second-order terms. A local contribution to the global cost is then computed in each variable neuron as
\begin{equation}
    \mathcal{C}_i^{(t+1)} = x_i^{(t)}\left[z_i^{(t+1)}+q_{ii}\right],
\end{equation}
obtaining the term in square brackets in \hyperref[eq:qubo-cost-decomposition]{equation \eqref{eq:qubo-cost-decomposition}}. An additional neuron, called \textbf{cost integrator}, is then required to sum over all the local contributions and obtain the total cost of the candidate solution $\mathbf{x}^{(t)}$.  Specifically, the neuron variables are connected to the cost integrator through synapses with unit weight and, spiking the contributions $\mathcal{C}_i^{(t+1)}$, accumulate in the cost integrator at step $t+2$ the quantity
\begin{align}
    \mathcal{C}^{(t+2)} & = \mathbf{1}^T\mathbf{C}^{(t+1)} \\
    & = \sum_{i=0}^{n-1} \mathcal{C}_i^{(t+1)},
\end{align}
which equals the decomposition of the QUBO cost obtained above. Hence, the proposed architecture encodes a QUBO problem and, given a candidate solution, can compute its cost in two steps.

\subsection*{Parallel variable updates}

The inherent parallelism of the QUBO SNN architecture can be exploited to evaluate the Metropolis criterion for all possible moves with unit Hamming distance. For ease of notation, let's denote with $\mathop{\Delta\mathcal{C}_i}(\mathbf{x})$ the variation in total cost associated to flipping the state of $x_i$, i.e., changing it from $0$ to $1$ or vice-versa. It can be computed from \hyperref[eq:qubo-cost-decomposition]{equation \eqref{eq:qubo-cost-decomposition}}, considering only the elements containing $x_i$, as
\begin{align}
    \mathop{\Delta\mathcal{C}_i}(\mathbf{x}) & = \pm\left[q_{ii} + \sum_{j\not=i}x_jq_{ji} + \sum_{j\not=i}x_jq_{ij} \right] \\
    & = \pm\left[q_{ii} + \sum_{j\not=i}(q_{ji}+q_{ij})x_j\right]\\
    \intertext{and, exploiting the assumption of symmetry $q_{ij}=q_{ji}$, reduced to}
    & = \pm\left[q_{ii} + 2\sum_{j\not=i}q_{ij}x_j\right]
    \label{eq:parallel-switching-condition}
\end{align}
where the $\pm$ sign is chosen based on the direction of the change, positive for changing $x_i$ from $0$ to $1$ and negative otherwise. Each neuron can compute \hyperref[eq:parallel-switching-condition]{equation \eqref{eq:parallel-switching-condition}} purely based on information that is locally available in the memory of this neuron, thus the parallel processors implementing the neurons can independently and in parallel calculate the equation to the impact of flipping its state on the total cost in parallel as
\begin{equation}
    \Delta\mathcal{C}_i(\mathbf{x}^{(t)}) = \left(q_{ii}+2z_i^{(t+1)}\right)
\begin{cases}
    +1 & \text{if } x_i^{(t)}=0 \\
    -1  & \text{if } x_i^{(t)}=1
\end{cases}.
\label{eq:delta-cost-snn}
\end{equation}
Hence, each neuron can sample a random number from the uniform distribution $\mathop\mathcal{U}(0,1)$ and, following \hyperref[eq:boltzmann]{equation \eqref{eq:boltzmann}}, update its state with probability
\begin{equation}
    p_i = \min \left(\exp\left(-\frac{\Delta\mathcal{C}_i(\mathbf{x}^{(t)})}{T}\right), 1\right).
\end{equation}
Each neuron encodes a copy of the temperature $T$ at the beginning of the run and update it according to the same annealing schedule. This, again, allows the processors encoding the neurons to work independently on their own integrated memory and thus to update all neurons in parallel.

It is important to note at this point that the obtained neural dynamics is an approximation of the full SA procedure. While the Metropolis criterion is evaluated for moves involving a single binary change, multiple moves are potentially accepted for each step, making the change in total cost computed with \hyperref[eq:delta-cost-snn]{equation \eqref{eq:delta-cost-snn}} an approximation. As we shall see next, it is beneficial to introduce a strategy to control the level of parallelization (and therefore approximation), breaking the symmetry of the system.

\subsection*{Stochastic refractory periods}

Parallel simulated annealing raises the issue that if two or more variables fulfill the Boltzmann condition, their joint flipping can worsen the solution due to interaction terms $Q_{ij}$. Several solutions have been proposed to solve this issue \cite{gre90}. Fujitsu's parallel Digital Annealer \citep{mtm20}, for example, solves this by using parallelism solely to determine in parallel which of the variables are suitable to fulfill the Boltzmann condition, while flipping only a single one of them. 

To speed up the convergence, we also enable parallel updates of many neurons, which pushes the Boltzmann machine away from a near-equilibrium state where neurons flip one at a time. In this non-equilibrium Boltzmann machine (NEBM), we introduce a \textit{stochastic refractory period}, preventing neurons from repeatedly flipping variables in successive steps.
Specifically, after a variable neuron changes its state, from $0$ to $1$ or vice-versa, further changes are inhibited for a random number of iterations. 
This change in the neural dynamics results in a reduced number of simultaneous variables updates per step, thus addressing the reported issue.
Moreover, the distribution of duration of the refractory period can be tuned to explicitly control the level of parallelization, with longer refractory periods resulting in more sequential updates.

\subsection*{Loihi 2 implementation}

Since Loihi 2 is conceived as an accelerator for spiking neural networks, its instruction set is optimized for the most common operations in this domain. The instruction set is thus constrained, which requires careful re-design of algorithms. As a reward, the resulting Loihi 2 architecture enables orders of magnitude gains in terms of latency and energy requirements, compared to classical CPUs.

Each Loihi 2 chip \cite{ofr21} features 128 neurocores that can simulate up to 8192 neurons each in parallel. Each neurocore is equipped with its own integrated memory, so that each neuron is equipped with its own local memory that allows quick and efficient iterative updates of neuronal states. The neural dynamics can be defined by the user using flexible micro-code, although computationally expensive operations like exponentiation and division are not supported for performance reasons. In each time step, each neuron receives synaptic inputs, updates its memory states, and can send a 32bit graded spike via synapses of 8-bit weights to other neurons. For more general but less performant instructions, each Loihi 2 chip features also embedded CPUs.

Evaluating the Metropolis criterion detailed in \hyperref[eq:boltzmann]{equation \eqref{eq:boltzmann}} on Loihi 2 neuro-cores presents different challenges. 
In particular, given the missing support for exponentiation and division, the inequality cannot be computed directly. Moreover, the condition needs to be evaluated at each step for all the variable neurons, making it inefficient to delegate its execution to an embedded CPUs. 
For this reason, we introduce an integer approximation to evaluate the stochastic switching condition.

At each step, given an estimated change in cost $\Delta\mathcal{C}$, the temperature level $T$, and a random number $r\in[0,1]$, the change in the variable state is accepted if and only if
\begin{equation}
    \exp \left(-\frac{\Delta\mathcal{C}}{T}\right) \geq r.
\end{equation}
Each Loihi 2 neuron has access to a new a pseudo-random number in each step. The generator produces integer values  $\texttt{rand}\in[0,2^{24}-1]$ which can be used in the modified switching condition
\begin{equation}
    \exp \left(-\frac{\Delta\mathcal{C}}{T}\right) \geq \frac{\texttt{rand}}{2^{24}-1}.
\end{equation}
Changing the exponentiation to base 2, which can be evaluated on Loihi, we obtain
\begin{equation}
    \exp_2 \left(-\frac{\Delta\mathcal{C}}{\hat{T}}\right) \geq \frac{\texttt{rand}}{2^{24}-1}
    \label{step}
\end{equation}
where the previous temperature $T$ is substituted with the change of variable
\begin{equation}
    \hat{T} = \frac{T}{\log_2 e}.
\end{equation}
Applying the logarithm in base $2$ on both sides of \hyperref[step]{equation \eqref{step}}, we obtain
\begin{align}
    -\frac{\Delta\mathcal{C}}{\hat{T}} &\geq \log_2\left(\frac{\texttt{rand}}{2^{24}-1}\right) \label{eq:approx}
    \intertext{which can be approximated as}
    \frac{\Delta\mathcal{C}}{\hat{T}} &\mathop< 24 - \log_2 \texttt{rand}.
    \intertext{The condition can be further simplified using the \emph{count leading zero} (clz) function. This operation, which is supported by Loihi 2, counts the number of left-most zeros in the binary representation of a number, which approximates the logarithm in base 2. In particular, given the 24-bit representation of \texttt{rand}, the inequality can be expressed as}
    \frac{\Delta\mathcal{C}}{\hat{T}} &\mathop< \mathop\text{clz}{(\texttt{rand})}
    \intertext{or, equivalently, if the temperature is non-zero}
    \Delta\mathcal{C} &\mathop< \hat{T}\mathop\text{clz}{(\texttt{rand})}
\end{align}
which doesn't contain exponentiation or division operations.

Hence, the Metropolis criterion can be approximately evaluated on Loihi 2 neuro-cores. 
In particular, if the cost is decreased (i.e., $\Delta\mathcal{C}\mathop<0$) or the current \texttt{rand} is equal to zero, the transition is automatically accepted.
Otherwise, \hyperref[eq:approx]{equation \eqref{eq:approx}} provides a cheap fixed-precision approximation of the Boltzmann ratio.

\section*{Benchmarking methods}
We analyzed the performance of our neuromorphic SA algorithm as an optimization solver for QUBO problems on a preliminary version of the NeuroBench benchmark \cite{yik_neurobench_2024}. The benchmark features a set of QUBO worklodas that search for the maximum independent sets of graphs. The goal of our benchmarking was to understand how the proposed solution compares to existing solutions on CPU, in terms of quality of the solutions, latency, energy consumption and scalability. 

When benchmarking optimization algorithms, two common approaches are typically possible: one defines a target solution quality and measures the time taken by different solvers to achieve it, while the other sets a target run-time and evaluates the quality of the solutions obtained. In our experiments, we opted for the latter approach.
At any given timeout, the quality of the solution is evaluated as the \emph{percentage gap} from the best known solution (BKS) of the associated instance:
\begin{equation}
    \text{gap}_{\text{\%}}(\mathbf{x}) = 100\left|\frac{\min(\mathcal{C}(\mathbf{x}),0)-\mathcal{C}(\mathbf{x}_{\text{BKS}})}{\mathcal{C}(\mathbf{x}_{\text{BKS}})}\right|
    \label{eq:gap}
\end{equation}
where the cost $\mathop\mathcal{C}(\mathbf{x})$ is truncated with a zero upper-bound. 
This choice intends to address the fact that some solvers report a best cost of $0$, associated to the trivial all-zeros solutions, when no better solution is found.
Hence, the metric can only assume values in $[0,100]$, with a value of $0\%$ corresponding to a solution as good as the best known one, and a value of $100\%$ corresponding to the worst case of the all-zero solutions.

We adopted two different CPU solvers for the comparison: the SA solver and the TS solver implemented in the D-Wave Samplers \texttt{v1.1.0} library \cite{dwavesamplers}. 
The library was compiled on Ubuntu \texttt{20.04.6 LTS} with GCC \texttt{9.4.0}  and Python \texttt{3.8.10}.
All measurements were obtained on a machine with Intel Core i9-7920X CPU @ 2.90 GHz
\footnote{The CPU has the following caches: L1i and L1d with 384KiB, L2 with 12 MiB, and L3 with 16.5 MiB.}
 and 128GB of DDR4 RAM, using Intel SoC Watch for Linux OS \texttt{2023.2.0}. 
Access to the code of our Loihi 2 solver used in these experiments is provided through the Intel Neuromorphic Research Community\footnote{\href{https://intel-ncl.atlassian.net/wiki/spaces/INRC/pages/1784807425/Join+the+INRC}{https://intel-ncl.atlassian.net/wiki/spaces/INRC/pages/1784807425/Join+the+INRC}}. 
The neuromorphic SA was executed on a Kapoho Point board using a single Loihi 2 chip, with Lava \texttt{0.8.0} and Lava Optimization \texttt{0.3.0}.

\subsection*{Maximum independent sets as benchmarking data set}
\label{section-mis}
Given an undirected graph $\mathcal{G}=(\mathcal{V}, \mathcal{E})$, an \emph{independent set} $\mathcal{I}$ is a subset of $\mathcal{V}$ such that, for any two vertices $u, v \in \mathcal{I}$, there is no edge connecting them. 
The Maximum Independent Set (MIS) problem consists in finding an independent set with maximum cardinality. It has been shown in the literature that the MIS problem is strongly NP-hard for a variety of graph structures \cite{punnen2022quadratic}. MIS has been applied to solve many industrial problems, e.g., the distribution of frequencies to 5G or WiFi access points without interference \cite{zww17}, the design of error-correcting code \cite{bps02} or the design of fault-tolerant semiconductor chips with redundant communication vias \cite{kut06}.

The MIS problem has a natural QUBO formulation: for each node $u\in\mathcal{V}$ in the graph, a binary variable $x_u$ is introduced to model the inclusion or not of $u$ in the candidate solution. Summing the quadratic terms $x_u^2$ will thus result in the size of the set of selected nodes. To penalize the selection of nodes that are not mutually independent, a penalization term is associated to the interactions $x_ux_v$ if $u$ and $v$ are connected. The resulting $\mathbf{Q}$ matrix coefficients are defined as
\begin{equation}
    q_{uv} = \begin{cases}
        -1 &\text{if }u=v\\
        \lambda &\text{if }u\not=v\text{ and }(u,v)\in\mathcal{E} \\
        0 & \text{otherwise}
    \end{cases}
    \label{eq:mis-qubo}
\end{equation}
where $\lambda>0$ is a large penalization term. 

\subsection*{Instances}
\label{prima}

We benchmarked on a set of randomly-generated MIS instances from the NeuroBench data set \cite{yik_neurobench_2024}: given a number of nodes $n$, a density value $d$ and a random seed $s$, the \texttt{MISProblem} generator produces an adjacency matrix. 
The associated QUBO formulation is then obtained based on \hyperref[eq:mis-qubo]{equation \eqref{eq:mis-qubo}}. 
As reported in \hyperref[tab:mis]{table \ref{tab:mis}}, we considered sizes from $10$ to $10^3$ and edge densities from $5$\% to $30$\%, for a total of $105$ instances.

In order to obtain the BKS, we adopted two different strategies depending on problem size.
The instances with up to $250$ nodes were solved using a branch-and-bound method from Gurobi \cite{gurobi2021gurobi}, obtaining the optimal solutions. 
However, solving to optimality larger instances would have required substantially more time and computational power\footnote{In a preliminary test, solving instances with $500$ variables took more than $6$ hours on our machine.}, and was thus beyond the scope of this thesis.
Hence, for instances with $500$ or more variables, we obtained the BKS executing the TABU solver on CPU with a timeout of $600$ s.

\begin{table}[t]
    \centering
    \caption{\label{tab:mis}Parameters used to generate the MIS instances.}
    \begin{tabular}{lc}
        \toprule
        \textbf{Parameter} & \textbf{Values} \\
        \midrule
        Nodes & 10, 25, 50, 100, 250, 500, 1000 \\
        Edge density & 5\%, 15\%, 30\% \\
        Random seed & 0, 1, 2, 3, 4 \\
        \bottomrule
    \end{tabular}
\end{table}

\section*{Benchmarking results}
We applied the three solvers on the MIS set of instances for different timeout values, from  $10^{-3}$ s to $10^{3}$ s, repeating each experiment with five initial points and random seeds.

\subsection*{Quality of the solutions}

\begin{figure}[t]
    \centering
    \begin{subfigure}{0.48\textwidth}
        \includegraphics[width=\linewidth]{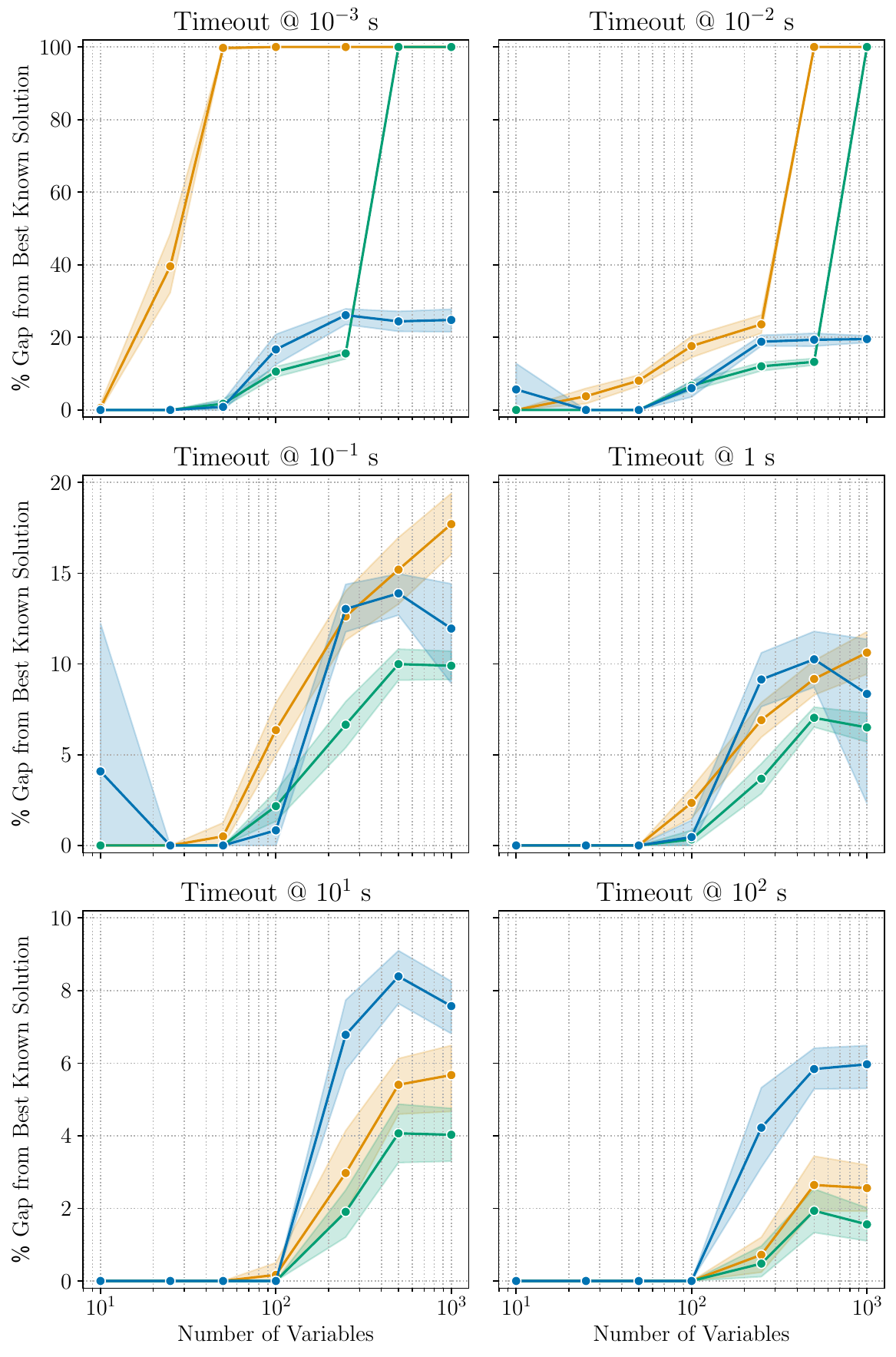}
    \vspace*{0.2cm}
    \end{subfigure}\\
    \begin{subfigure}{0.3\textwidth}
        \includegraphics[width=\linewidth]{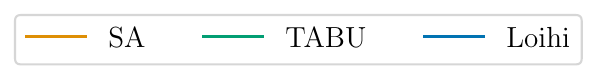}
    \end{subfigure}
    \caption{Percentage gap from the best known solution for the MIS instances ($15\%$ density), for increasing timeout values. }
    \label{fig:mis-optimality-015}
\end{figure}

\hyperref[fig:mis-optimality-015]{Figure \ref{fig:mis-optimality-015}} reports the results on solution quality for the instances with 15\% edge density, as measured by the percentage gap from the BKS.

The relative performance of the solvers can be categorized in three different regimes. 
For tight time constraints, at $10^{-2}$ s timeout or less, the CPU solvers struggle to produce good solution for large instances.
In particular, our neuromorphic solver provides feasible solutions for instances up to $40\times$ larger compared to SA, and $4\times$ larger compared to TABU. 
At intermediate timeout values, the three solvers demonstrate similar performance (with a small advantage for TABU), all producing solutions with a percentage gap lower than 20\% across all problem sizes.
For longer run times, lasting $10$ s or more, TABU provides the lowest percentage gap, followed by SA on CPU and our proposed solution. 
However, the differences in percentage gap between the three solvers are quite limited and keep reducing with increased timeouts, suggesting that all three algorithmic approaches would eventually reach the same level of optimality.

Our results in \hyperref[fig:mis-optimality-015]{figure \ref{fig:mis-optimality-015}} present an unexpected behaviour. 
In particular, for some combinations of solvers and timeouts, instances with $1000$ variables were solved with a better percentage gap compared to instances with $500$ variables, while the percentage gap is expected to monotonically increase with problem size. 
A possible explanation for this phenomenon is the methodology adopted to obtain the BKS. 
While instances up to $250$ variables were solved to optimality, the BKS for $500$ and $1000$ variables were obtained with TABU, with a long and fixed timeout of $600$ s. 
Hence, we hypothesize that the smaller instances, with $500$ variables, have a stronger BKS compared to the larger ones with $1000$ variables, resulting in the observed anomaly of the percentage gap. 
Further investigation would be required to fully assess this behaviour, for example obtaining the optimal solutions also for these larger problem sizes.

\begin{figure}[t]
    \centering
        \includegraphics[width=\linewidth]{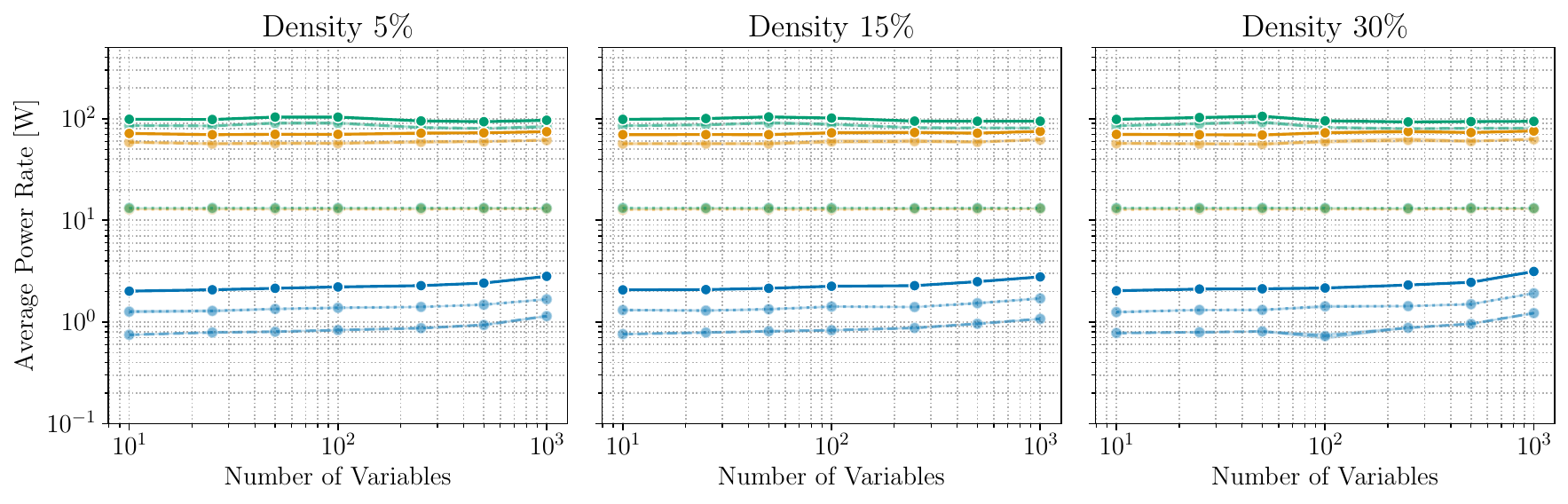}
    \vspace*{0.5cm}
        \includegraphics[width=0.3\linewidth]{Figures/legend.pdf}
        \caption{\label{fig:resources-015} Power consumption running the MIS instances with the different solvers. The CPU solvers require up to $37\times$ more power than our neuromorphic SA on Loihi 2.}
\end{figure}

\subsection*{Power consumption}

To evaluate the energy efficiency of neuromorphic hardware compared to baseline algorithms, we profiled the power consumption of the three solvers on the full set of MIS instances. Since MIS workloads were run for a fixed timeout, differences in power consumption are equivalent to differences in energy consumption of the chips. 
The main results are reported in \hyperref[fig:resources-015]{figure \ref{fig:resources-015}}.

Our neuromorphic SA running on Loihi 2 has an average total power of $2.62$ W, while the SA and TABU solvers reach respectively $87.78$ W and $97.56$ W. 
Hence, the proposed solution results in $33.5\times$ lower energy consumption compared to SA on CPU, and $37.24\times$ lower compared to TABU.
Moreover, the plots show how the energy consumption of Loihi slightly increases with problem size, while the CPU solvers have a constant power overall. 
This trend can be explained by the fact that, while Loihi can exploit multiple cores based on problem size, the CPU solvers are limited to a single core. 

Overall, the energy consumption results, coupled with the competitive quality of the solutions, strongly supports the proposed solver as a more energy-efficient approach to QUBO.

\section*{Discussion}
This paper details the architecture, implementation, and performance of our hardware-aware, fine-grained parallel simulated annealing algorithm on Loihi 2 for solving QUBO problems.
We observe that the structure of combinatorial optimization algorithms is well-suited to neuromorphic hardware architectures like Loihi 2. Our solver exploited this synergy to find feasible solutions to QUBO problems in as little as 1 ms, and with up to $37\times$ lower energy consumption than a state-of-the-art Tabu solver on CPU. Continuing work on the Loihi 2-aware algorithm, advanced partitioning on the Loihi 2 chip, and scaling to multi-chip systems promises to further improve the time to solution, energy consumption, and optimality of the solver.

A variety of well-known problems can be formulated as QUBO, often exhibiting a significant degree of unstructured sparsity well-suited to our approach \cite{glover2022applications}. For commercial applications with size, weight, and power (SWaP) constraints, such as those faced in edge computing contexts, our approach could enable significant benefits compared to state-of-the-art methods. For example, in mobile robotics, a fast, energy efficient QUBO solver could support routing, path planning \cite{clark2019roboQUBO}, and robotic scheduling \cite{leib2023optimization} with lower latency, longer battery life, and the capacity to handle more complex scenarios. 
Many industrial applications running on high-performance computing could also benefit from significantly faster QUBO solvers. In finance for example, this could include problems such as portfolio optimization \cite{phb21}, high-frequency arbitrage trading \cite{zcw22}, and credit-risk assessment \cite{mhh23}.

The presented algorithm and preliminary results leave several limitations to be addressed in future research.
First, the present study provides results for problems up to 1000 variables, and we have further verified the capacity to scale to roughly 4000 variables on a single Loihi 2 chip. But in its current implementation, the solver is not capable of scaling up to very large-scale problems (e.g. up to 1M variables) due to limitations in the synaptic encoding and transmission of messages in multi-chip networks on Loihi 2. Future work will investigate algorithmic and software solutions to achieve state-of-the-art problem size, such as problem decomposition.
Second, these initial results focus on maximum independent set problems due to the ability to arbitrarily scale and sparsify these synthetic problems. In additional testing, we observe that many other QUBO problems possess significantly greater complexity than MIS, as expressed in the roughness of the QUBO cost landscape and in the number of acceptably near-optimal solutions in the search space. Additional research should extend these initial benchmarks to a broader representation of real-world applications to properly understand the performance benefits of our approach.
Third, some established QUBO formulations, such as the Traveling Salesman Problem, require a numerical precision that is unachievable with our current implementation of the algorithm, which is restricted to 8-bit integer quadratic cost coefficients. For such problems, the solver could combine multiple Loihi 2 synapses or leverage scaled numerical representations to achieve higher dynamic range. Finally, future work should evaluate the performance of our algorithm against QUBO-specific accelerators, such as those based on FPGAs \cite{tatsumura_fpga-based_2019,goto_combinatorial_2019,tatsumura_scaling_2021,goto_high-performance_2021, kashimata_efficient_2024}, application-specific CMOS \cite{aramon_physics-inspired_2019, yoshimura2020cmos}, analogue hardware \cite{mmb22, jsh23}, and D-Wave's Quantum Annealer \cite{jag11, oot19}. 

In conclusion, our new solver for NP-hard discrete QUBO, together with our previous solver for polynomial-time, continuous convex quadratic programming \citep{mfr24}, demonstrates the performance of the Intel Loihi 2 neuromorphic processor for mathematical optimization.
This approach could find useful applications in SWaP-constrained edge computing as well as larger datacenter systems.
\label{sec:discussion}

\bibliography{./Lib/bibliography}

\begin{thebibliography}{10}
\expandafter\ifx\csname url\endcsname\relax
  \def\url#1{\texttt{#1}}\fi
\expandafter\ifx\csname urlprefix\endcsname\relax\def\urlprefix{URL }\fi
\providecommand{\bibinfo}[2]{#2}
\providecommand{\eprint}[2][]{\url{#2}}

\bibitem{mfr24}
\bibinfo{author}{Mangalore, A.~R.}, \bibinfo{author}{Fonseca, G.~A.}, \bibinfo{author}{Risbud, S.~R.}, \bibinfo{author}{Stratmann, P.} \& \bibinfo{author}{Wild, A.}
\newblock \bibinfo{title}{Neuromorphic quadratic programming for efficient and scalable model predictive control: Towards advancing speed and energy efficiency in robotic control}.
\newblock \emph{\bibinfo{journal}{IEEE Robotics \& Automation Magazine}} \bibinfo{pages}{2--12} (\bibinfo{year}{2024}).

\bibitem{davies_advancing_2021}
\bibinfo{author}{Davies, M.} \emph{et~al.}
\newblock \bibinfo{title}{Advancing {Neuromorphic} {Computing} {With} {Loihi}: {A} {Survey} of {Results} and {Outlook}}.
\newblock \emph{\bibinfo{journal}{Proceedings of the IEEE}} \textbf{\bibinfo{volume}{109}}, \bibinfo{pages}{911--934} (\bibinfo{year}{2021}).
\newblock \urlprefix\url{https://ieeexplore.ieee.org/document/9395703/}.

\bibitem{FonsecaGuerra2020Stochastic}
\bibinfo{author}{Guerra, G.~F.}
\newblock \emph{\bibinfo{title}{Stochastic Processes for Neuromorphic Hardware}}.
\newblock Ph.D. thesis, \bibinfo{school}{University of Manchester} (\bibinfo{year}{2020}).

\bibitem{and14}
\bibinfo{author}{Lucas, A.}
\newblock \bibinfo{title}{Ising formulations of many np problems}.
\newblock \emph{\bibinfo{journal}{Frontiers in Physics}} \textbf{\bibinfo{volume}{2}} (\bibinfo{year}{2014}).
\newblock \urlprefix\url{https://www.frontiersin.org/journals/physics/articles/10.3389/fphy.2014.00005}.

\bibitem{Barahona1982}
\bibinfo{author}{Barahona, F.}
\newblock \bibinfo{title}{On the computational complexity of ising spin glass models}.
\newblock \emph{\bibinfo{journal}{Journal of Physics A: Mathematical and General}} \textbf{\bibinfo{volume}{15}}, \bibinfo{pages}{3241} (\bibinfo{year}{1982}).

\bibitem{DunningEtAl2018}
\bibinfo{author}{Dunning, I.}, \bibinfo{author}{Gupta, S.} \& \bibinfo{author}{Silberholz, J.}
\newblock \bibinfo{title}{What works best when? a systematic evaluation of heuristics for max-cut and qubo}.
\newblock \emph{\bibinfo{journal}{{INFORMS} Journal on Computing}} \textbf{\bibinfo{volume}{30}} (\bibinfo{year}{2018}).

\bibitem{glover1986future}
\bibinfo{author}{Glover, F.}
\newblock \bibinfo{title}{Future paths for integer programming and links to artificial intelligence}.
\newblock \emph{\bibinfo{journal}{Computers \& operations research}} \textbf{\bibinfo{volume}{13}}, \bibinfo{pages}{533--549} (\bibinfo{year}{1986}).

\bibitem{dwavesamplers}
\bibinfo{author}{{D-Wave Systems Inc.}}
\newblock \emph{\bibinfo{title}{D-Wave Samplers software package}} (\bibinfo{year}{2018}).

\bibitem{optBySimAnn}
\bibinfo{author}{Kirkpatrick, S.}, \bibinfo{author}{Gelatt, C.~D.} \& \bibinfo{author}{Vecchi, M.~P.}
\newblock \bibinfo{title}{Optimization by simulated annealing}.
\newblock \emph{\bibinfo{journal}{Science}} \textbf{\bibinfo{volume}{220}}, \bibinfo{pages}{671--680} (\bibinfo{year}{1983}).
\newblock \urlprefix\url{https://www.science.org/doi/abs/10.1126/science.220.4598.671}.
\newblock \eprint{https://www.science.org/doi/pdf/10.1126/science.220.4598.671}.

\bibitem{gre90}
\bibinfo{author}{Greening, D.~R.}
\newblock \bibinfo{title}{Parallel simulated annealing techniques}.
\newblock \emph{\bibinfo{journal}{Physica D: Nonlinear Phenomena}} \textbf{\bibinfo{volume}{42}}, \bibinfo{pages}{293--306} (\bibinfo{year}{1990}).
\newblock \urlprefix\url{https://www.sciencedirect.com/science/article/pii/0167278990900843}.

\bibitem{carpenter2010graphics}
\bibinfo{author}{Carpenter, J.} \& \bibinfo{author}{Wilkinson, T.~D.}
\newblock \bibinfo{title}{Graphics processing unit--accelerated holography by simulated annealing}.
\newblock \emph{\bibinfo{journal}{Optical Engineering}} \textbf{\bibinfo{volume}{49}}, \bibinfo{pages}{095801--095801} (\bibinfo{year}{2010}).

\bibitem{oshiyama_benchmark_2022}
\bibinfo{author}{Oshiyama, H.} \& \bibinfo{author}{Ohzeki, M.}
\newblock \bibinfo{title}{Benchmark of quantum-inspired heuristic solvers for quadratic unconstrained binary optimization}.
\newblock \emph{\bibinfo{journal}{Scientific Reports}} \textbf{\bibinfo{volume}{12}}, \bibinfo{pages}{2146} (\bibinfo{year}{2022}).
\newblock \urlprefix\url{https://www.nature.com/articles/s41598-022-06070-5}.

\bibitem{goto_combinatorial_2019}
\bibinfo{author}{Goto, H.}, \bibinfo{author}{Tatsumura, K.} \& \bibinfo{author}{Dixon, A.~R.}
\newblock \bibinfo{title}{Combinatorial optimization by simulating adiabatic bifurcations in nonlinear {Hamiltonian} systems}.
\newblock \emph{\bibinfo{journal}{Science Advances}} \textbf{\bibinfo{volume}{5}}, \bibinfo{pages}{eaav2372} (\bibinfo{year}{2019}).
\newblock \urlprefix\url{https://www.science.org/doi/10.1126/sciadv.aav2372}.

\bibitem{glover2022applications}
\bibinfo{author}{Glover, F.}, \bibinfo{author}{Kochenberger, G.} \& \bibinfo{author}{Du, Y.}
\newblock \bibinfo{title}{Applications and computational advances for solving the qubo model}.
\newblock In \emph{\bibinfo{booktitle}{The Quadratic Unconstrained Binary Optimization Problem: Theory, Algorithms, and Applications}}, \bibinfo{pages}{39--56} (\bibinfo{publisher}{Springer}, \bibinfo{year}{2022}).

\bibitem{tatsumura_fpga-based_2019}
\bibinfo{author}{Tatsumura, K.}, \bibinfo{author}{Dixon, A.~R.} \& \bibinfo{author}{Goto, H.}
\newblock \bibinfo{title}{{FPGA}-{Based} {Simulated} {Bifurcation} {Machine}}.
\newblock In \emph{\bibinfo{booktitle}{2019 29th {International} {Conference} on {Field} {Programmable} {Logic} and {Applications} ({FPL})}}, \bibinfo{pages}{59--66} (\bibinfo{publisher}{IEEE}, \bibinfo{address}{Barcelona, Spain}, \bibinfo{year}{2019}).
\newblock \urlprefix\url{https://ieeexplore.ieee.org/document/8892209/}.

\bibitem{tatsumura_scaling_2021}
\bibinfo{author}{Tatsumura, K.}, \bibinfo{author}{Yamasaki, M.} \& \bibinfo{author}{Goto, H.}
\newblock \bibinfo{title}{Scaling out {Ising} machines using a multi-chip architecture for simulated bifurcation}.
\newblock \emph{\bibinfo{journal}{Nature Electronics}} \textbf{\bibinfo{volume}{4}}, \bibinfo{pages}{208--217} (\bibinfo{year}{2021}).
\newblock \urlprefix\url{https://www.nature.com/articles/s41928-021-00546-4}.

\bibitem{goto_high-performance_2021}
\bibinfo{author}{Goto, H.} \emph{et~al.}
\newblock \bibinfo{title}{High-performance combinatorial optimization based on classical mechanics}.
\newblock \emph{\bibinfo{journal}{Science Advances}} \textbf{\bibinfo{volume}{7}}, \bibinfo{pages}{eabe7953} (\bibinfo{year}{2021}).
\newblock \urlprefix\url{https://www.science.org/doi/10.1126/sciadv.abe7953}.

\bibitem{kashimata_efficient_2024}
\bibinfo{author}{Kashimata, T.}, \bibinfo{author}{Yamasaki, M.}, \bibinfo{author}{Hidaka, R.} \& \bibinfo{author}{Tatsumura, K.}
\newblock \bibinfo{title}{Efficient and {Scalable} {Architecture} for {Multiple}-{Chip} {Implementation} of {Simulated} {Bifurcation} {Machines}}.
\newblock \emph{\bibinfo{journal}{IEEE Access}} \textbf{\bibinfo{volume}{12}}, \bibinfo{pages}{36606--36621} (\bibinfo{year}{2024}).
\newblock \urlprefix\url{https://ieeexplore.ieee.org/document/10460551/}.

\bibitem{aramon_physics-inspired_2019}
\bibinfo{author}{Aramon, M.} \emph{et~al.}
\newblock \bibinfo{title}{Physics-{Inspired} {Optimization} for {Quadratic} {Unconstrained} {Problems} {Using} a {Digital} {Annealer}}.
\newblock \emph{\bibinfo{journal}{Frontiers in Physics}} \textbf{\bibinfo{volume}{7}}, \bibinfo{pages}{48} (\bibinfo{year}{2019}).
\newblock \urlprefix\url{https://www.frontiersin.org/article/10.3389/fphy.2019.00048/full}.

\bibitem{yoshimura2020cmos}
\bibinfo{author}{Yoshimura, C.}, \bibinfo{author}{Hayashi, M.}, \bibinfo{author}{Takemoto, T.} \& \bibinfo{author}{Yamaoka, M.}
\newblock \bibinfo{title}{Cmos annealing machine: A domain-specific architecture for combinatorial optimization problem}.
\newblock In \emph{\bibinfo{booktitle}{2020 25th Asia and South Pacific Design Automation Conference (ASP-DAC)}}, \bibinfo{pages}{673--678} (\bibinfo{organization}{IEEE}, \bibinfo{year}{2020}).

\bibitem{mmb22}
\bibinfo{author}{Mohseni, N.}, \bibinfo{author}{McMahon, P.~L.} \& \bibinfo{author}{Byrnes, T.}
\newblock \bibinfo{title}{Ising machines as hardware solvers of combinatorial optimization problems}.
\newblock \emph{\bibinfo{journal}{Nature Reviews Physics}} \textbf{\bibinfo{volume}{4}}, \bibinfo{pages}{363--379} (\bibinfo{year}{2022}).

\bibitem{jsh23}
\bibinfo{author}{Jiang, M.}, \bibinfo{author}{Shan, K.}, \bibinfo{author}{He, C.} \& \bibinfo{author}{Li, C.}
\newblock \bibinfo{title}{Efficient combinatorial optimization by quantum-inspired parallel annealing in analogue memristor crossbar}.
\newblock \emph{\bibinfo{journal}{Nature Communications}} \textbf{\bibinfo{volume}{14}}, \bibinfo{pages}{5927} (\bibinfo{year}{2023}).

\bibitem{jag11}
\bibinfo{author}{Johnson, M.~W.} \emph{et~al.}
\newblock \bibinfo{title}{Quantum annealing with manufactured spins}.
\newblock \emph{\bibinfo{journal}{Nature}} \textbf{\bibinfo{volume}{473}}, \bibinfo{pages}{194--198} (\bibinfo{year}{2011}).

\bibitem{oot19}
\bibinfo{author}{Okada, S.}, \bibinfo{author}{Ohzeki, M.}, \bibinfo{author}{Terabe, M.} \& \bibinfo{author}{Taguchi, S.}
\newblock \bibinfo{title}{Improving solutions by embedding larger subproblems in a d-wave quantum annealer}.
\newblock \emph{\bibinfo{journal}{Scientific reports}} \textbf{\bibinfo{volume}{9}}, \bibinfo{pages}{2098} (\bibinfo{year}{2019}).

\bibitem{steidel2018mapcoloring}
\bibinfo{author}{Steidel, J.}
\newblock \emph{\bibinfo{title}{Solving Map Coloring Problems on Analog Neuromorphic Hardware}}.
\newblock \bibinfo{type}{Bachelorarbeit}, \bibinfo{school}{Universit\"{a}t Heidelberg} (\bibinfo{year}{2018}).

\bibitem{Corder2018VertexCover}
\bibinfo{author}{Corder, K.}, \bibinfo{author}{Monaco, J.~V.} \& \bibinfo{author}{Vindiola, M.~M.}
\newblock \bibinfo{title}{Solving vertex cover via ising model on a neuromorphic processor}.
\newblock In \emph{\bibinfo{booktitle}{2018 IEEE International Symposium on Circuits and Systems (ISCAS)}}, \bibinfo{pages}{1--5} (\bibinfo{publisher}{IEEE}, \bibinfo{year}{2018}).
\newblock \urlprefix\url{https://doi.org/10.1109/ISCAS.2018.8351248}.

\bibitem{Rahman2017Cognitive}
\bibinfo{author}{Rahman, N.}, \bibinfo{author}{Atahary, T.}, \bibinfo{author}{Taha, T.~M.} \& \bibinfo{author}{Douglass, S.}
\newblock \bibinfo{title}{Cognitive domain ontologies on the truenorth neurosynaptic system}.
\newblock In \emph{\bibinfo{booktitle}{Proceedings of the International Joint Conference on Neural Networks (IJCNN)}}, \bibinfo{pages}{3543--3550} (\bibinfo{publisher}{IEEE}, \bibinfo{year}{2017}).
\newblock \urlprefix\url{https://doi.org/10.1109/IJCNN.2017.7966302}.

\bibitem{dwo21}
\bibinfo{author}{Davies, M.} \emph{et~al.}
\newblock \bibinfo{title}{Advancing neuromorphic computing with loihi: A survey of results and outlook}.
\newblock \emph{\bibinfo{journal}{Proceedings of the IEEE}} \textbf{\bibinfo{volume}{109}}, \bibinfo{pages}{911--934} (\bibinfo{year}{2021}).

\bibitem{Binas2016Spiking}
\bibinfo{author}{Binas, J.}, \bibinfo{author}{Indiveri, G.} \& \bibinfo{author}{Pfeiffer, M.}
\newblock \bibinfo{title}{Spiking analog vlsi neuron assemblies as constraint satisfaction problem solvers}.
\newblock In \emph{\bibinfo{booktitle}{2016 IEEE International Symposium on Circuits and Systems (ISCAS)}}, \bibinfo{pages}{2094--2097} (\bibinfo{publisher}{IEEE}, \bibinfo{year}{2016}).
\newblock \urlprefix\url{https://doi.org/10.1109/ISCAS.2016.7539013}.

\bibitem{Mniszewski2019Graph}
\bibinfo{author}{Mniszewski, S.~M.}
\newblock \bibinfo{title}{Graph partitioning as quadratic unconstrained binary optimization (qubo) on spiking neuromorphic hardware}.
\newblock In \emph{\bibinfo{booktitle}{Proceedings of the International Conference on Neuromorphic Systems}} (\bibinfo{publisher}{ACM}, \bibinfo{year}{2019}).
\newblock \urlprefix\url{https://ouci.dntb.gov.ua/en/works/45zDRVOl/}.

\bibitem{henke_sampling_2023}
\bibinfo{author}{Henke, K.}, \bibinfo{author}{Pelofske, E.}, \bibinfo{author}{Hahn, G.} \& \bibinfo{author}{Kenyon, G.}
\newblock \bibinfo{title}{Sampling binary sparse coding {QUBO} models using a spiking neuromorphic processor}.
\newblock In \emph{\bibinfo{booktitle}{Proceedings of the 2023 {International} {Conference} on {Neuromorphic} {Systems}}}, \bibinfo{pages}{1--5} (\bibinfo{publisher}{ACM}, \bibinfo{address}{Santa Fe NM USA}, \bibinfo{year}{2023}).
\newblock \urlprefix\url{https://dl.acm.org/doi/10.1145/3589737.3606003}.

\bibitem{Alom2017QUBO}
\bibinfo{author}{Alom, M.~Z.}, \bibinfo{author}{Essen, B.~V.}, \bibinfo{author}{Moody, A.~T.}, \bibinfo{author}{Widemann, D.~P.} \& \bibinfo{author}{Taha, T.~M.}
\newblock \bibinfo{title}{Quadratic unconstrained binary optimization (qubo) on neuromorphic computing system}.
\newblock In \emph{\bibinfo{booktitle}{Proceedings of the International Joint Conference on Neural Networks (IJCNN)}}, \bibinfo{pages}{3922--3929} (\bibinfo{publisher}{IEEE}, \bibinfo{year}{2017}).
\newblock \urlprefix\url{https://doi.org/10.1109/IJCNN.2017.7966350}.

\bibitem{Liang2019Neuromorphic}
\bibinfo{author}{Liang, D.} \& \bibinfo{author}{Indiveri, G.}
\newblock \bibinfo{title}{A neuromorphic computational primitive for robust context-dependent decision making and context-dependent stochastic computation}.
\newblock \emph{\bibinfo{journal}{IEEE Transactions on Circuits and Systems II: Express Briefs}} \textbf{\bibinfo{volume}{66}}, \bibinfo{pages}{843--847} (\bibinfo{year}{2019}).
\newblock \urlprefix\url{https://doi.org/10.1109/TCSII.2018.2889084}.

\bibitem{fonseca_guerra_using_2017}
\bibinfo{author}{Fonseca~Guerra, G.~A.} \& \bibinfo{author}{Furber, S.~B.}
\newblock \bibinfo{title}{Using {Stochastic} {Spiking} {Neural} {Networks} on {SpiNNaker} to {Solve} {Constraint} {Satisfaction} {Problems}}.
\newblock \emph{\bibinfo{journal}{Frontiers in Neuroscience}} \textbf{\bibinfo{volume}{11}}, \bibinfo{pages}{714} (\bibinfo{year}{2017}).
\newblock \urlprefix\url{http://journal.frontiersin.org/article/10.3389/fnins.2017.00714/full}.

\bibitem{Henke2023Sampling}
\bibinfo{author}{Henke, K.}, \bibinfo{author}{Pelofske, E.}, \bibinfo{author}{Hahn, G.} \& \bibinfo{author}{Kenyon, G.~T.}
\newblock \bibinfo{title}{Sampling binary sparse coding qubo models using a spiking neuromorphic processor}.
\newblock In \emph{\bibinfo{booktitle}{Proceedings of the 2023 International Conference on Neuromorphic Systems (ICONS)}} (\bibinfo{publisher}{ACM}, \bibinfo{address}{Santa Fe, NM, USA}, \bibinfo{year}{2023}).
\newblock \urlprefix\url{https://doi.org/10.48550/ARXIV.2306.01940}.

\bibitem{mtm20}
\bibinfo{author}{Matsubara, S.} \emph{et~al.}
\newblock \bibinfo{title}{Digital annealer for high-speed solving of combinatorial optimization problems and its applications}.
\newblock In \emph{\bibinfo{booktitle}{2020 25th Asia and South Pacific Design Automation Conference (ASP-DAC)}}, \bibinfo{pages}{667--672} (\bibinfo{year}{2020}).

\bibitem{ofr21}
\bibinfo{author}{Orchard, G.} \emph{et~al.}
\newblock \bibinfo{title}{Efficient neuromorphic signal processing with loihi 2}.
\newblock In \emph{\bibinfo{booktitle}{2021 IEEE Workshop on Signal Processing Systems (SiPS)}}, \bibinfo{pages}{254--259} (\bibinfo{year}{2021}).

\bibitem{yik_neurobench_2024}
\bibinfo{author}{Yik, J.} \emph{et~al.}
\newblock \bibinfo{title}{{NeuroBench}: {A} {Framework} for {Benchmarking} {Neuromorphic} {Computing} {Algorithms} and {Systems}} (\bibinfo{year}{2024}).
\newblock \urlprefix\url{http://arxiv.org/abs/2304.04640}.
\newblock \bibinfo{note}{ArXiv:2304.04640 [cs]}.

\bibitem{punnen2022quadratic}
\bibinfo{author}{Punnen, A.~P.}
\newblock \emph{\bibinfo{title}{The quadratic unconstrained binary optimization problem}} (\bibinfo{publisher}{Springer}, \bibinfo{year}{2022}).

\bibitem{zww17}
\bibinfo{author}{Zhou, J.}, \bibinfo{author}{Wang, L.}, \bibinfo{author}{Wang, W.} \& \bibinfo{author}{Zhou, Q.}
\newblock \bibinfo{title}{Efficient graph-based resource allocation scheme using maximal independent set for randomly- deployed small star networks}.
\newblock \emph{\bibinfo{journal}{Sensors}} \textbf{\bibinfo{volume}{17}} (\bibinfo{year}{2017}).
\newblock \urlprefix\url{https://www.mdpi.com/1424-8220/17/11/2553}.

\bibitem{bps02}
\bibinfo{author}{Butenko, S.}, \bibinfo{author}{Pardalos, P.}, \bibinfo{author}{Sergienko, I.}, \bibinfo{author}{Shylo, V.} \& \bibinfo{author}{Stetsyuk, P.}
\newblock \bibinfo{title}{Finding maximum independent sets in graphs arising from coding theory}.
\newblock In \emph{\bibinfo{booktitle}{Proceedings of the 2002 ACM Symposium on Applied Computing}}, SAC '02, \bibinfo{pages}{542–546} (\bibinfo{publisher}{Association for Computing Machinery}, \bibinfo{address}{New York, NY, USA}, \bibinfo{year}{2002}).
\newblock \urlprefix\url{https://doi.org/10.1145/508791.508897}.

\bibitem{kut06}
\bibinfo{author}{Lee, K.-Y.} \& \bibinfo{author}{Wang, T.-C.}
\newblock \bibinfo{title}{Post-routing redundant via insertion for yield/reliability improvement}.
\newblock In \emph{\bibinfo{booktitle}{Asia and South Pacific Conference on Design Automation, 2006.}}, \bibinfo{pages}{6 pp.--} (\bibinfo{year}{2006}).

\bibitem{gurobi2021gurobi}
\bibinfo{author}{Gurobi~Optimization, L.}
\newblock \bibinfo{title}{Gurobi optimizer reference manual} (\bibinfo{year}{2021}).

\bibitem{clark2019roboQUBO}
\bibinfo{author}{Clark, J.} \emph{et~al.}
\newblock \bibinfo{title}{Towards real time multi-robot routing using quantum computing technologies}.
\newblock In \emph{\bibinfo{booktitle}{Proceedings of the international conference on high performance computing in Asia-Pacific Region}}, \bibinfo{pages}{111--119} (\bibinfo{year}{2019}).

\bibitem{leib2023optimization}
\bibinfo{author}{Leib, D.} \emph{et~al.}
\newblock \bibinfo{title}{An optimization case study for solving a transport robot scheduling problem on quantum-hybrid and quantum-inspired hardware}.
\newblock \emph{\bibinfo{journal}{Scientific Reports}} \textbf{\bibinfo{volume}{13}}, \bibinfo{pages}{18743} (\bibinfo{year}{2023}).

\bibitem{phb21}
\bibinfo{author}{Phillipson, F.} \& \bibinfo{author}{Bhatia, H.~S.}
\newblock \bibinfo{title}{Portfolio optimisation using the d-wave quantum annealer}.
\newblock In \bibinfo{editor}{Paszynski, M.}, \bibinfo{editor}{Kranzlm{\"u}ller, D.}, \bibinfo{editor}{Krzhizhanovskaya, V.~V.}, \bibinfo{editor}{Dongarra, J.~J.} \& \bibinfo{editor}{Sloot, P. M.~A.} (eds.) \emph{\bibinfo{booktitle}{Computational Science -- ICCS 2021}}, \bibinfo{pages}{45--59} (\bibinfo{publisher}{Springer International Publishing}, \bibinfo{address}{Cham}, \bibinfo{year}{2021}).

\bibitem{zcw22}
\bibinfo{author}{Zhuang, X.-N.}, \bibinfo{author}{Chen, Z.-Y.}, \bibinfo{author}{Wu, Y.-C.} \& \bibinfo{author}{Guo, G.-P.}
\newblock \bibinfo{title}{Quantum computational quantitative trading: high-frequency statistical arbitrage algorithm}.
\newblock \emph{\bibinfo{journal}{New Journal of Physics}} \textbf{\bibinfo{volume}{24}}, \bibinfo{pages}{073036} (\bibinfo{year}{2022}).
\newblock \urlprefix\url{https://dx.doi.org/10.1088/1367-2630/ac7f26}.

\bibitem{mhh23}
\bibinfo{author}{Ma, F.}, \bibinfo{author}{He, Y.} \& \bibinfo{author}{Hu, J.}
\newblock \bibinfo{title}{Application of qubo model in credit score card combination optimization}.
\newblock \emph{\bibinfo{journal}{Highlights in Science, Engineering and Technology}} \textbf{\bibinfo{volume}{68}}, \bibinfo{pages}{304–312} (\bibinfo{year}{2023}).
\newblock \urlprefix\url{https://drpress.org/ojs/index.php/HSET/article/view/12092}.

\end{thebibliography}

\section*{Acknowledgments}
The authors are indebt to the their colleagues at the Neuromorphic Computing Lab for discussions and support. Particular thanks go to Sumit Bam Shrestha for discussions on efficient micro-code implementations. 

\section*{Author Contributions}
PS developed the algorithm. AP, GG, and SR developed the software. AP and GG designed the experiments. AP performed the experiments and the analysis. PS, AP, GG, SR wrote the manuscript. All authors discussed the results and critically revised the manuscript.

\section*{Competing Interests}

The authors declare no competing financial interests.

\end{document}